\pdfoutput=1
\documentclass[11pt]{article}

\usepackage[preprint]{acl}

\usepackage{times}
\usepackage{latexsym}

\usepackage[utf8]{inputenc}

\usepackage{microtype}

\usepackage{inconsolata}

\usepackage[T1]{fontenc}    % use 8-bit T1 fonts
\usepackage{url}            % simple URL typesetting
\usepackage{booktabs}       % professional-quality tables
\usepackage{amsfonts}       % blackboard math symbols
\usepackage{array}
\usepackage{longtable}
\usepackage{wrapfig}
\usepackage{lscape}
\usepackage{adjustbox}
\usepackage{multirow}
\usepackage{tcolorbox}
\usepackage{pgfplots}
\usepackage{pgfplotstable}
\pgfplotsset{compat=1.18}
\usepackage[normalem]{ulem}
\useunder{\uline}{\ul}{}
\usepackage{enumitem}
\usepackage{cleveref}
\usepackage[symbol]{footmisc}
\usepackage{fvextra}
\usepackage{listings}
\usepackage{graphicx}

\definecolor{darkgreen}{RGB}{0,100,0}  % Dark green color
\definecolor{darkred}{RGB}{139,0,0}    % Dark red color

\lstdefinestyle{wrappedverbatim}{
  basicstyle=\small\ttfamily,
  breaklines=true,
  breakatwhitespace=false,
  columns=flexible,
  keepspaces=true,
  showstringspaces=false,
  frame=none,
  backgroundcolor=\color{white!95!gray}
}

\title{Auto-Cypher: Improving LLMs on Cypher generation via LLM-supervised generation-verification framework}

\author{
    Aman Tiwari$^{*}$, 
    Shiva Krishna Reddy Malay\thanks{Co-first authors with equal contribution.}, 
    Vikas Yadav, \\ 
    \textbf{Masoud Hashemi}, 
    \textbf{Sathwik Tejaswi Madhusudhan} \\[0.5em]
    ServiceNow \\[0.5em]
    \texttt{\{aman.tiwari, shivakrishnareddy.ma, vikas.yadav,} \\ 
    \texttt{masoud.hashemi, sathwikt.madhusudhan\}@servicenow.com}
}

\begin{document}
\maketitle

\renewcommand{\thefootnote}{}
\footnotetext{The dataset used in this work is available at: \url{https://huggingface.co/datasets/ServiceNow-AI/SynthCypher}.}
\renewcommand{\thefootnote}{\arabic{footnote}}

\begin{abstract}

Graph databases like Neo4j are gaining popularity for handling complex, interconnected data, over traditional relational databases in modeling and querying relationships. While translating natural language into SQL queries is well-researched, generating Cypher queries for Neo4j remains relatively underexplored. In this work, we present an automated, LLM-Supervised, pipeline to generate high-quality synthetic data for Text2Cypher. Our Cypher data generation pipeline introduces LLM-As-Database-Filler, a novel strategy for ensuring Cypher query correctness, thus resulting in high quality generations. Using our pipeline, we generate high quality Text2Cypher data - \texttt{SynthCypher} containing 29.8k instances across various domains and queries with varying complexities. Training open-source LLMs like LLaMa-3.1-8B, Mistral-7B, and QWEN-7B on \texttt{SynthCypher} results in performance gains of up to 40\% on the Text2Cypher test split and 30\% on the SPIDER benchmark, adapted for graph databases.

\end{abstract}

\section{Introduction}

As the use of graph databases like Neo4j \cite{neo4j} grows, converting natural language into Cypher queries (Text2Cypher) is becoming increasingly important. Cypher \cite{francis2018cypher}, designed for querying and analyzing graph data, is well-suited for applications such as social networks, recommendation systems, and knowledge graphs \cite{ji2021survey}. However, generating Cypher queries from natural language poses challenges due to the complexity of graph structures, which surpasses that of relational databases. Large language models (LLMs) have shown promise in Text2Cypher tasks. However, unlike Text2SQL, which benefits from extensive datasets and benchmarks \cite{Deng_2021, li2023llmservedatabaseinterface, shi2024surveyemployinglargelanguage}, resources for training LLMs to generate accurate Cypher queries are limited.

\begin{figure}
    \centering
    \includegraphics[width=\linewidth]{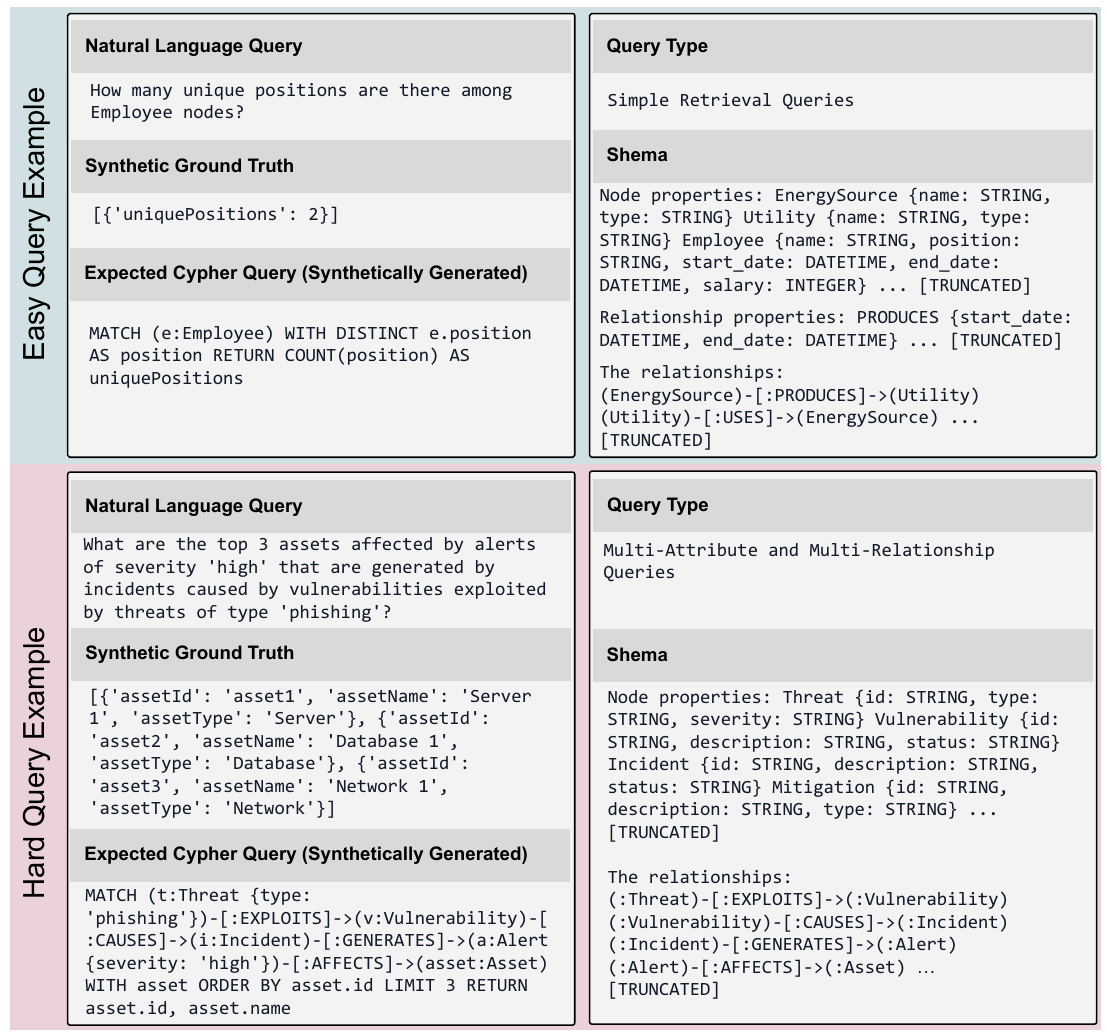}\vspace{-2mm}
    \caption{\footnotesize{Example figure showing input \textit{Natural Language Query} which is converted to \textit{Cypher Query} for the given \textit{Schema}. The example on top shows an easy retrieval question while bottom example shows complex Multi-Attribute and Multi-Relationship Query.}}
    \vspace{-7mm}
    \label{fig:enter-label}
\end{figure}

%, making automation difficult.

To address these limitations, we introduce an automated data generation pipeline specifically designed for Text2Cypher tasks. Our proposed pipeline generates high-quality synthetic Cypher queries to enable supervised fine-tuning of LLMs for Text2Cypher task, ensuring more precise natural language to Cypher translation. The pipeline begins by generating diverse graphical schemas across a wide range of domains and complexity. For these schemas, we generate natural language questions covering substantial taxonomies (such as simple retrieval, complex aggregation, path-finding, etc.), which are then used to create corresponding Cypher queries. A key feature of our pipeline is the LLM-As-Database-Filler which generates synthetic Neo4j databases. Finally, in the validation step only executable queries that produce correct results are retained. This results in \texttt{SynthCypher}, a robust and diverse dataset for Text2Cypher tasks.

Using \texttt{SynthCypher}, we trained LLMs including Qwen 2.5~\cite{hui2024qwen2}, Llama 3.1~\cite{maaten2024llama3}, and Mistral~\cite{jiang2023mistral}, along with their code-specialized versions. % Having been trained on a larger volume of code-related data, these code-specific models were expected to offer improved performance in generating Cypher queries due to their enhanced understanding of syntax, structure, and execution semantics. However, our results did not show a significant advantage for code-specialized models over the general-purpose ones. 
Moreover, due to the lack of a widely accepted benchmark for Cypher, we adapted the SPIDER~\cite{deng2020structure} benchmark, originally designed for Text2SQL, to serve as a benchmark for graph databases.

%(1) we present a pipeline that ensures valid Cypher queries through robust validation steps. (2) We train a state-of-the-art LLM specifically for Text2Cypher. (3) We perform extensive analysis across multiple datasets, demonstrating the model's effectiveness and generalizability. Notably, we obtain 40\% improvement in accuracy on the state of the art 7B \& 8B models on the proposed Cypher benchmark and 30\% on the modified SPIDER benchmark.

% Our contributions include: 
% (1) We present a Cypher code generation pipeline that ensures valid Cypher queries through robust validation steps, resulting in a high-quality dataset, \texttt{SynthCypher}, with 29.8k training and 2k test samples, covering over 109 query taxonomies and 700 domains. The novel data validation method within our pipeline, LLM-As-Database-Filler, generates synthetic Neo4j databases that enable execution of Cypher queries for correctness.  
% (2) To show effectiviness of \texttt{SynthCypher}, we fine-tune state-of-the-art LLMs, including Qwen, Llama 3.1, and Mistral, for text2Cypher tasks. Our extensive empirical evaluations show that models finetuned with \texttt{SynthCypher} achieve up to a 40\% improvement in accuracy on the \texttt{SynthCypher} test split using 7B \& 8B models and 30\% on the modified SPIDER benchmark for graph databases.
% (3) We adapt the SPIDER benchmark (originally designed for Text2SQL) to evaluate Cypher query generation, addressing lack of Cypher benchmarks. 

Our contributions are threefold: (1) We introduce a pipeline for Cypher code generation that ensures valid queries via robust validation, producing the high-quality dataset \texttt{SynthCypher} with 29.8k training and 2k test samples, covering 109 query taxonomies and 700 domains. The LLM-As-Database-Filler method generates synthetic Neo4j databases to verify query correctness. (2) We fine-tune state-of-the-art LLMs (Qwen, Llama 3.1, Mistral) on text2Cypher tasks. Models fine-tuned with \texttt{SynthCypher} show up to 40\% accuracy improvement on 7B \& 8B models and 30\% on a modified SPIDER benchmark. (3) We adapt the SPIDER benchmark for Cypher query generation, addressing the lack of Cypher benchmarks.

\section{Related Work}

\paragraph{}Prior works~\cite{liang2021querying,hains2023natural} on natural language querying of knowledge graphs using Cypher has mostly focused on traditional NER based extraction approaches which make them both limited in scope, and cumbersome to write. LLMs have shown promising potential for Text2Cypher task where recently, Neo4j Labs published a gpt-4o generated dataset~\cite{tomasjno}, initiating first efforts on Text2Cypher data generation. Importantly, this Text2Cypher data without any validation steps on a limited domain set, with only 6 query types on HuggingFace~\cite{wolf2019huggingface} results only in 50\% correctly executable cyphers. Concurrent (peer-reviewed unpublished) Synth2C \cite{zhong2024synthet2cgeneratingsyntheticdata} 
%is the latest work on synthetic data generation for Cypher which 
generates Cyphers using Gpt-4o similar to Neo4j Labs as well as a templatized pipeline with traditional NLP techniques and llm-as-judge to validate generated cypher descriptions against original questions. However, this technique again does not check for execution correctness and is furthermore limited only to Medical domain (with datasets not publicly available).

% Text2SQL has been comparatively well studied in literature with abundance of benchmarks and datasets; SPIDER \cite{deng2020structure} especially is a prominent dataset that covers a wide range of real world scenarios. However, real-world applicability remains an open question as evidenced by SPIDER-V2 \cite{cao2024spider2} benchmark with GPT-4 still at 6\% pass@1.

% Text2SQL has been well studied\textcolor{red}{CITE}\, with benchmarks like SPIDER \cite{deng2020structure} covering diverse real-world scenarios. Due to the lack of Text2Cypher benchmarks, we convert SPIDER into a Text2Cypher benchmark, transforming its tables into Neo4j databases for our experiments. This enables the evaluation of LLMs on Cypher query generation in graph databases, addressing a critical gap in Text2Cypher evaluation.

% Write a passage about SPIDER and how/why it is important to evaluate on that. Sell the point that our numbers are better than many SQL-LLM solutions, hence Tex2Cypher research domain is very important. Mention this in Introduction 2nd or 3rd passage also. 

% Write a short passage on open source, code instruct models and how they compare with OpenAI GPT models. Why do we need open-sourced code-LLMs and why we focus on them in this paper. 

\begin{figure*}[t!]
    \centering
    \includegraphics[width=0.9\linewidth]{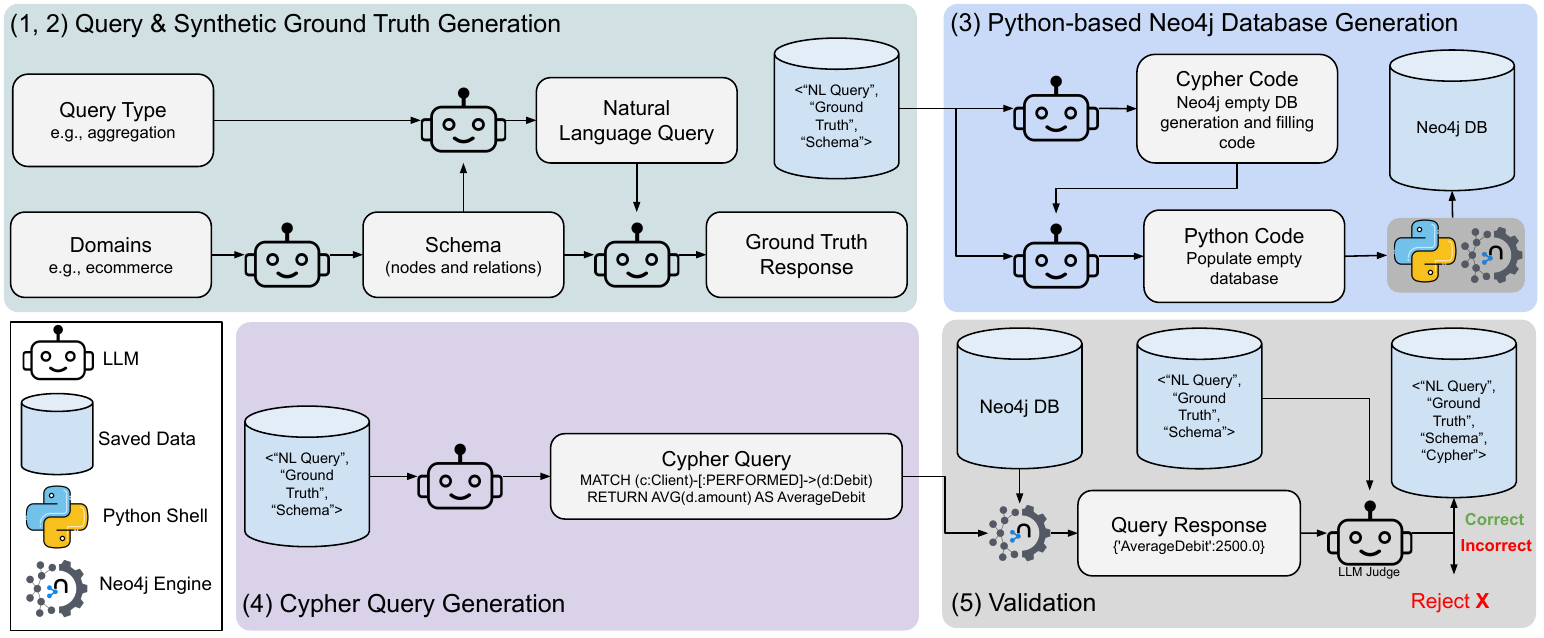}\vspace{-3mm}
    \caption{\footnotesize{Overview of the \texttt{SynthCypher} data generation pipeline, illustrating domain and schema creation, query and ground truth generation, database population, Cypher query generation, and validation steps.}}
    \label{fig:complete_pipeline}
    \vspace{-3mm}
\end{figure*}

\section{Data Generation Pipeline}

Synthetic data generation \cite{xu2023wizardlm,luo2023wizardcoder,ouyang2022training} have proven highly effective.% at mimicking or surpassing human-generated data. 
We use LLMs such as Llama 3.1 70B \cite{maaten2024llama3}, Mixtral 8x22B \cite{Jiang2024MixtralOE}, and GPT-4 \cite{openai2023gpt4} to automatically generate diverse domains, schemas, natural language queries, and Cypher queries. Our pipeline covers a broad range of domains and query types, ensuring diversity across topics and difficulty. %It is fully automated, requiring no human intervention. 
From data generation to validation, all steps are autonomously managed by models and scripts, allowing the process to run at scale. Generated Cypher queries are executed and validated against expected results to ensure quality. 

\noindent\textbf{\emph{Step 1: Schema Generation}}: We begin by random selection of the seed domains (e.g., e-commerce, inventory management) from Neo4j \cite{neo4j} example databases. We then use Mixtral to expand these domains to cover 700 distinct domains. A skeleton schema is generated for each domain, outlining the nodes and relationships (Block 1 in Figure~\ref{fig:complete_pipeline}). These schemas are validated with GPT-4 for correctness and manually reviewed for coherence and real-world utility in 25\% of cases. See Appendix~\ref{apdx_schema_generation} for more details on schema generation.

\noindent\textbf{\emph{Step 2: Natural Language Question and Ground Truth Generation} }For each schema, we generate questions based on 109 predefined query types, such as ``Simple Retrieval`` or ``Sub-Graph Queries'' (Block 2 in Figure~\ref{fig:complete_pipeline}). % Note that these query types are also \textit{Evol}-ed from a seed list of query types. 
 A dummy ground truth answer for each query is also generated. %, representing the expected output in response to the question. 
  In the next stage, we fill the database with entries including this dummy answer as the right answer for the question. See Appendix~\ref{apdx_question_generation} for further details on question generation and query types.

\noindent\textbf{\emph{Step 3: Neo4j Database Population}} An empty Neo4j database for each question is created which is populated with synthetic data that fits the schema, question, and ground truth. Python-based code, generated by GPT-4, is used to create and populate the database with nodes, relationships, and data, ensuring consistency between the schema and ground truth (Block 3 in Figure~\ref{fig:complete_pipeline}). To the best of our knowledge, this strategy of filling the database conditioned on a arbitrarily chosen dummy ground truth has not been explored in literature before. %, and maybe applicable in non-Cypher paradigms also. 
 Reverse filling the database in this way enables execution of Cypher queries to check for execution success and Cypher-code correctness. Appendix~\ref{apdx_database_infill}(\ref{fig:code_plan_generation},\ref{fig:python_code_generation}) provides more details on the data population process.

\noindent\textbf{\emph{Step 4: Cypher Query Generation}} Next, the LLM generates Cypher queries for each question (Block 4 in Figure~\ref{fig:complete_pipeline}). Following latest work in inference time scaling, we allow the LLM to amply reason through various aspects of the Cypher query, such as relevant nodes, relationships, properties, nuances of the question as well as best coding practices. This iterative chain-of-thought reasoning process coupled with execution checks against the synthetically filled database ensures only the highest quality data is generated. See Appendix~\ref{apdx_cypher_gen} for details on query generation.

\noindent\textbf{\emph{Step 5: Validation of Cypher Queries}} To ensure accuracy, we validate the generated Cypher queries by executing them on the synthetic Neo4j database from Step-3 (Block 5 in Figure~\ref{fig:complete_pipeline}). The results are compared to the expected ground truth, and only queries that return correct results are retained and others retried upto 5 times before discarding. GPT-4 is used as a judge to validate (Prompt~\ref{fig:execution_match} in appendix.) the retrieved data against the ground-truth and ensure correctness. %, with additional manual validation performed on a sample of queries. 

At the end of this process, we have a high-quality dataset, \texttt{SynthCypher}, that includes schema, Neo4j database, natural language questions, Cypher queries, and execution results. This dataset can be used for training and evaluating models aimed at converting natural language into Cypher code.

% \begin{table*}[h!]
%     \centering
%     \footnotesize
%     \begin{tabular}{l|ccc|ccc}
%         \toprule
%         & \multicolumn{3}{c}{\textbf{Ours}} & \multicolumn{3}{c}{\textbf{Neo4j-Labs (Text2Cypher GPT-o)}} \\ 
%         \cmidrule(lr){2-4} \cmidrule(lr){5-7}
%         \textbf{Split} & \textbf{Count} & \textbf{Schema} & \textbf{Validation}  & \textbf{Count} & \textbf{Schema} & \textbf{Validation} \\ 
%         \midrule
%         Train & 29,838 & 528 & \textcolor{darkgreen}{\checkmark} & 7,735 & 15 & \textcolor{darkred}{\texttimes} \\ 
%         Test  & 2,000  & 165 & \textcolor{darkgreen}{\checkmark} & -     & -  & -          \\ 
%         \bottomrule
%     \end{tabular}\vspace{-2mm}
%     \caption{Dataset split and count summary in comparison to Neo4J-Labs' Text2Cypher.}
%     \label{tab:dataset_table}
%     \vspace{-5mm}
% \end{table*}
\vspace{-2mm}
\begin{table}[h!]

\small  % This will make the table font smaller
\begin{tabular}{p{1cm}p{1cm}p{1cm}p{1cm}p{2cm}}
\toprule
\textbf{Split} & \textbf{Dataset} & \textbf{Count} & \textbf{Schema} & \textbf{Validation} \\ 
\midrule
Train          & Ours       & 29,838 & 528    & \checkmark \\ 
Train          & Neo4j & 7,735  & 15     & \texttimes \\ 
& Labs &   &      & \\ 
\midrule
Test           & Ours       & 2,000  & 165    & \checkmark \\ 
Test           & Neo4j & -      & -      & -          \\ 
& Labs &   &      & \\ 
\bottomrule
\end{tabular}
\vspace{-2mm}
\caption{\footnotesize Comparison of datasets across training and testing splits}
\vspace{-7mm}
\label{tab:dataset_table}
\end{table}

\section{Experimental Setup}
\label{sec:experimental_setup}
\paragraph{Data Setup}: We used our dataset consisting of 25.8k samples spanning 109 query types and 528 schemas (Table~\ref{tab:dataset_table}) for training. The 109 query types in our \texttt{SynthCypher} represent diverse real-world Cypher use cases. For testing, we employed a separate dataset of 4k samples, covering all 109 query types across 165 schemas not included in train. This split ensures that the model is evaluated on a broad range of query complexities and schema variations. As an additional test dataset, we also adapt the popular SPIDER-SQL~\cite{yu-etal-2018-spider} for Text2Cypher by modeling each table as a node and foreign key relationships.\footnote{Junction tables where all columns are foreign keys are still modeled as nodes for ease of data filling.}

% \begin{table*}[h!]
%     \centering
%     \begin{tabular}{|c|ccc|ccc|}
%         \hline
%         & \multicolumn{3}{c|}{\textbf{Ours}} & \multicolumn{3}{c|}{\textbf{Neo4J-Labs (Text2Cypher GPT-o)}} \\
%         \textbf{Split} & \textbf{Count} & \textbf{Schema} & \textbf{Validation}  & \textbf{Count} & \textbf{Schema} & \textbf{Validation} \\ \hline
%         Train & 29838 & 528 & \emoji{check-mark-button} & 7740 & 15 & \emoji{cross-mark} \\ \hline
%         Test & 2000 & 165 & \emoji{check-mark-button} & - & - & -\\ \hline
%     \end{tabular}
%     \caption{Dataset split and count summary in comparison to the only existing text2Cypher }
%     \label{tab:dataset_table}
% \end{table*}

\begin{table}[ht]
\centering
\footnotesize
\begin{tabular}{llcc}
\toprule
\textbf{Setup} & \textbf{Model} & \textbf{SynCy-test} & \textbf{SPIDER} \\ 
% \midrule
% \multirow{5}{*}{\rotatebox[origin=c]{90}{Baseline}} & T5-SR &  & 72.4\\
% & RESDSQL-3B + NatSQL &  & 72.0\\
% & T5-3B + PICARD &  & 71.9 \\	
% & SADGA + GAP &  & 70.1 \\
\midrule
\multirow{5}{*}{\rotatebox[origin=c]{90}{Base IFT}} & Llama-3.1-8B & 30.9 & 30.8\\
& Mistral v0.2 7B & 31.1 & 38.3\\
& Qwen2-7B & 14.6 & 16.6 \\	
& Code-Llama-7B & 38.5 & 37.3 \\
& Code-Qwen-2.5 & 50.85 & 57.3\\
\midrule
\multirow{5}{*}{\rotatebox[origin=c]{90}{Instruct}} & Llama-3.1-8B & 40.2 & 37.9\\
& Mistral v0.2 7B & 27.7 & 25.2\\
& Qwen2-7B & 29.2 & 33.5\\	
& Code-Llama-7B & 34.0 & 32.8\\
& Code-Qwen-2.5 & 29.2 & 50.8\\
\midrule
\multirow{5}{*}[1.25em]{\rotatebox[origin=c]{90}{
  \parbox{2.2cm}{\centering {\tiny Base + SynCy} \\ (Ours)}
}} & Llama-3.1-8B & \textbf{71.4} & \textbf{62.2}\\
& Mistral v0.2 7B & 69.4 & 61.3\\
& Qwen2-7B & 67.1 & 55.2\\	
& Code-Llama-7B & 67.1 & 61.2\\
& Code-Qwen-2.5 & 70.1 & 62.1\\

\bottomrule
\end{tabular}
\vspace{-2mm}
\caption{Last block shows Funetuning when our SynthCypher SFT data is mixed with MagiCoder}
\label{tab:Result_table}
\vspace{-5mm}
\end{table}

\paragraph{Experiment Setup}: %Due to lack of prior published work for this problem, 
We begin our experimentation by analysing the capabilities of the current state of the art 7B/8B models on Text2Cypher. %To this end, we investigate the extent to which pre-training contributes to the performance, as well as the impact of large scale Supervised\-Fine\-Tuning (SFT).
 We initially fully finetune three general base models, i.e. Llama 3.1 model\cite{maaten2024llama3}, Mistral-v0.2-7B\cite{jiang2023mistral} and Qwen-2-7B\cite{hui2024qwen2}, along with two code based models CodeLlama-7B and QwenCoder-2.5-7B. We use UltraChat-200K\cite{ding2023enhancing} for instruction-finetuning (IFT) the general models and MagiCoder-117K\cite{luo2023wizardcoder} for finetuning code models. These instruction finetuned model would highlight effectiveness of existing IFT datasets on Text2Cypher task. Next, we also benchmark off-the-shelf instruct versions of these models on both SynthCypher and SPIDER-Cypher. 

\begin{figure}
    \centering
    \includegraphics[width=0.85\linewidth]{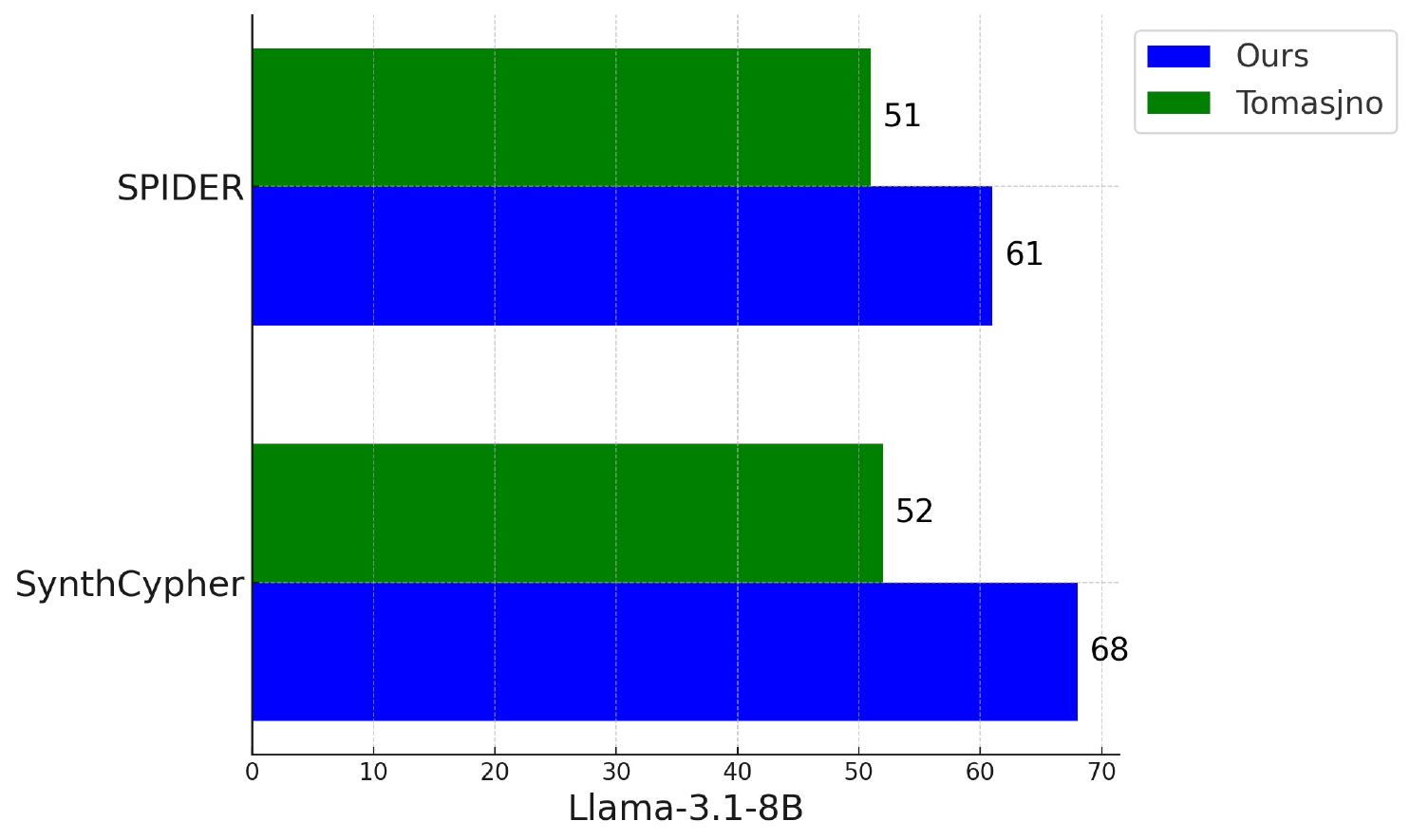}
    \vspace{-2mm}
    \caption{\footnotesize Evaluation on SynthCypher and SPIDER test splits from Llama3.1-8B fine-tuned with equal train size of down-sampled SynthCypher (ours) data and Neo4j Text2Cypher data. %\footnote{https://huggingface.co/tomasonjo/text2cypher-demo-16bit} 
    }
    \label{fig:bar_chart_comparison}
    \vspace{-5mm}
\end{figure}

\noindent In our last setup, we concatenate our generated \texttt{SynthCypher} data with UltraChat for finetuning the general LLMs (LLaMa and Mistral) and with MagiCoder for finetuning the code LLMs (CodeLLaMa and QwenCoder). We use learning rate=1e-05, batch size of 128 over three epochs for training and take the best one based on a sub-sampled validation set. To the best of our knowledge, there is only one other dataset for this task, i.e. Tomasnjo\_gpt4o\cite{tomasjno} which is a created by naively prompting GPT-4o and checking only the cypher produces \textit{some} results. The authors indicate that only 50\% of the cypher passed the test cases on a small (27 samples) human generated benchmark. We show comparison of Tomasnjo\_gpt4o with our subsampled \texttt{SynthCypher} data (to match the training size of 7.7K instances) in \cref{fig:bar_chart_comparison}. We chose our best performing base LLM (LLaMa-3.1-8B) for this comparison. % We perform an ablation study on this dataset on our best performing model Llama-3.1-8B, by retraining the base model to control for the difference in training data size and lack of chain\-of\-thought information.

\textbf{Metric:} We use an LLM-as-Judge version (prompt\ref{fig:execution_match}) of Exact Match where GPT-4o gives a score of 1 if all requested information in the question is present in the results from the execution, and 0 otherwise.

\section{Results}
As shown in Table-\ref{tab:Result_table}, our \texttt{SynthCypher} dataset leads to significant improvements on both benchmarks across models. We draw several key observations: 
\begin{enumerate}[label={\bf(\arabic*)},itemsep=0em,topsep=0em,  wide, labelwidth=!, labelindent=0pt]
\item {\bf Need of Text2Cypher datasets} - Both off-the-shelf instruct LLMs and our finetuned LLMs on base IFT datasets achieve very low performance. Thus, highlighting lack of Text2Cypher alignment of code LLMs and need of more Text2Cypher IFT datasets.
\item {\bf Effectiveness of \texttt{SynthCypher}} - LLMs finetuned with IFT data mix containing \texttt{SynthCypher} achieve 40\% absolute improvement over the base IFT datasets and 30\% over off-the-shelf instruct LLMs. These encouraging improvements highlight effectiveness of \texttt{SynthCypher} and directions for future works. 
\item {\bf \texttt{SynthCypher} pipeline} - Comparison shown in \cref{fig:bar_chart_comparison} clearly highlights effectiveness of our pipeline and \texttt{SynthCypher} over other existing dataset generated using GPT-4o. This highlights benefits of step-by-step controlled data generation for Text2Cypher.
\end{enumerate}

\section{Conclusion}
In this work, we highlight and address the Text2Cypher gap in current open source models, and introduced a novel pipeline to automatically generate and validate high quality Text2Cypher data. Our presented dataset \texttt{SynthCypher} from our pipeline leads to substantial performance improvements across multiple LLMs. We also provide two evaluation benchmarks for future works in this direction. %We hypothesize that Cypher can be an excellent tool for AI agents to reason over both closed and open domain knowledge graphs due to the inherently more semantic nature of graph relationships compared to foreign keys in a relational database. 
% This work represents an important first step in that direction.

\section{Limitations}
While synthetic data generation strategies have played a crucial role in open source LLM models, these strategies may pose risks in terms of reinforcing model biases, thereby resulting in a data distribution that may not model real world scenarios, or worse yet, cause real world harm (especially when applied to social graph networks). Furthermore, we have limited this research to smaller models and it is not clear if the same strategy would work on larger models.

SPIDER test dataset has been publicly released as of Feb-2024 and it is not clear if any of that data went into the pre-training of base models or the Instruct models we considered.

\bibliography{custom}

\begin{thebibliography}{22}
\providecommand{\natexlab}[1]{#1}

\bibitem[{tom(2024)}]{tomasjno}
 2024.
\newblock Huggingface: tomasonjo/text2cypher-gpt4o-clean.
\newblock \url{https://huggingface.co/datasets/tomasonjo/text2cypher-gpt4o-clean?row=0}.
\newblock Accessed: 2024-09-12.

\bibitem[{neo(2024)}]{neo4j}
 2024.
\newblock Neo4j.
\newblock \url{https://neo4j.com/}.
\newblock Accessed: 2024-09-12.

\bibitem[{Deng et~al.(2020)Deng, Awadallah, Meek, Polozov, Sun, and Richardson}]{deng2020structure}
Xiang Deng, Ahmed~Hassan Awadallah, Christopher Meek, Oleksandr Polozov, Huan Sun, and Matthew Richardson. 2020.
\newblock Structure-grounded pretraining for text-to-sql.
\newblock \emph{arXiv preprint arXiv:2010.12773}.

\bibitem[{Deng et~al.(2021)Deng, Awadallah, Meek, Polozov, Sun, and Richardson}]{Deng_2021}
Xiang Deng, Ahmed~Hassan Awadallah, Christopher Meek, Oleksandr Polozov, Huan Sun, and Matthew Richardson. 2021.
\newblock \href {https://doi.org/10.18653/v1/2021.naacl-main.105} {Structure-grounded pretraining for text-to-sql}.
\newblock In \emph{Proceedings of the 2021 Conference of the North American Chapter of the Association for Computational Linguistics: Human Language Technologies}. Association for Computational Linguistics.

\bibitem[{Ding et~al.(2023)Ding, Chen, Xu, Qin, Zheng, Hu, Liu, Sun, and Zhou}]{ding2023enhancing}
Ning Ding, Yulin Chen, Bokai Xu, Yujia Qin, Zhi Zheng, Shengding Hu, Zhiyuan Liu, Maosong Sun, and Bowen Zhou. 2023.
\newblock \href {https://arxiv.org/abs/2305.14233} {Enhancing chat language models by scaling high-quality instructional conversations}.
\newblock \emph{Preprint}, arXiv:2305.14233.

\bibitem[{Francis et~al.(2018)Francis, Green, Guagliardo, Libkin, Lindaaker, Marsault, Plantikow, Rydberg, Selmer, and Taylor}]{francis2018cypher}
Nadime Francis, Alastair Green, Paolo Guagliardo, Leonid Libkin, Tobias Lindaaker, Victor Marsault, Stefan Plantikow, Mats Rydberg, Petra Selmer, and Andr{\'e}s Taylor. 2018.
\newblock Cypher: An evolving query language for property graphs.
\newblock In \emph{Proceedings of the 2018 international conference on management of data}, pages 1433--1445.

\bibitem[{Hains et~al.(2023)Hains, Khmelevsky, and Tachon}]{hains2023natural}
Ga{\'e}tan J. D.~R. Hains, Youry Khmelevsky, and Thibaut Tachon. 2023.
\newblock From natural language to graph queries.
\newblock In \emph{2023 IEEE 19th International Conference on Software Engineering Research, Management and Applications (SERA)}, pages 1--6. IEEE.

\bibitem[{Hui et~al.(2024)Hui, Yang, Cui, Yang, Liu, Zhang, Liu, Zhang, Yu, Dang et~al.}]{hui2024qwen2}
Binyuan Hui, Jian Yang, Zeyu Cui, Jiaxi Yang, Dayiheng Liu, Lei Zhang, Tianyu Liu, Jiajun Zhang, Bowen Yu, Kai Dang, et~al. 2024.
\newblock Qwen2. 5-coder technical report.
\newblock \emph{arXiv preprint arXiv:2409.12186}.

\bibitem[{Ji et~al.(2021)Ji, Pan, Cambria, Marttinen, and Philip}]{ji2021survey}
Shaoxiong Ji, Shirui Pan, Erik Cambria, Pekka Marttinen, and S~Yu Philip. 2021.
\newblock A survey on knowledge graphs: Representation, acquisition, and applications.
\newblock \emph{IEEE transactions on neural networks and learning systems}, 33(2):494--514.

\bibitem[{Jiang et~al.(2023)Jiang, Sablayrolles, Mensch, Bamford, Chaplot, Casas, Bressand, Lengyel, Lample, Saulnier et~al.}]{jiang2023mistral}
Albert~Q Jiang, Alexandre Sablayrolles, Arthur Mensch, Chris Bamford, Devendra~Singh Chaplot, Diego de~las Casas, Florian Bressand, Gianna Lengyel, Guillaume Lample, Lucile Saulnier, et~al. 2023.
\newblock Mistral 7b.
\newblock \emph{arXiv preprint arXiv:2310.06825}.

\bibitem[{Jiang et~al.(2024)Jiang, Sablayrolles, Roux, Mensch, Savary, Bamford, Chaplot, de~Las~Casas, Hanna, Bressand, Lengyel, Bour, Lample, Lavaud, Saulnier, Lachaux, Stock, Subramanian, Yang, Antoniak, Scao, Gervet, Lavril, Wang, Lacroix, and Sayed}]{Jiang2024MixtralOE}
Albert~Q. Jiang, Alexandre Sablayrolles, Antoine Roux, Arthur Mensch, Blanche Savary, Chris Bamford, Devendra~Singh Chaplot, Diego de~Las~Casas, Emma~Bou Hanna, Florian Bressand, Gianna Lengyel, Guillaume Bour, Guillaume Lample, L'elio~Renard Lavaud, Lucile Saulnier, Marie-Anne Lachaux, Pierre Stock, Sandeep Subramanian, Sophia Yang, Szymon Antoniak, Teven~Le Scao, Th{\'e}ophile Gervet, Thibaut Lavril, Thomas Wang, Timoth{\'e}e Lacroix, and William~El Sayed. 2024.
\newblock Mixtral of experts.
\newblock \emph{ArXiv}, abs/2401.04088.

\bibitem[{Li et~al.(2023)Li, Hui, Qu, Yang, Li, Li, Wang, Qin, Cao, Geng, Huo, Zhou, Ma, Li, Chang, Huang, Cheng, and Li}]{li2023llmservedatabaseinterface}
Jinyang Li, Binyuan Hui, Ge~Qu, Jiaxi Yang, Binhua Li, Bowen Li, Bailin Wang, Bowen Qin, Rongyu Cao, Ruiying Geng, Nan Huo, Xuanhe Zhou, Chenhao Ma, Guoliang Li, Kevin C.~C. Chang, Fei Huang, Reynold Cheng, and Yongbin Li. 2023.
\newblock \href {https://arxiv.org/abs/2305.03111} {Can llm already serve as a database interface? a big bench for large-scale database grounded text-to-sqls}.
\newblock \emph{Preprint}, arXiv:2305.03111.

\bibitem[{Liang et~al.(2021)Liang, Stockinger, de~Farias, Anisimova, and Gil}]{liang2021querying}
Shiqi Liang, Kurt Stockinger, Tarcisio~Mendes de~Farias, Maria Anisimova, and Manuel Gil. 2021.
\newblock \href {https://doi.org/10.1186/s40537-020-00383-w} {Querying knowledge graphs in natural language}.
\newblock \emph{Journal of Big Data}, 8(1):3.

\bibitem[{Luo et~al.(2023)Luo, Xu, Zhao, Sun, Geng, Hu, Tao, Ma, Lin, and Jiang}]{luo2023wizardcoder}
Ziyang Luo, Can Xu, Pu~Zhao, Qingfeng Sun, Xiubo Geng, Wenxiang Hu, Chongyang Tao, Jing Ma, Qingwei Lin, and Daxin Jiang. 2023.
\newblock Wizardcoder: Empowering code large language models with evol-instruct.
\newblock \emph{arXiv preprint arXiv:2306.08568}.

\bibitem[{OpenAI(2023)}]{openai2023gpt4}
OpenAI. 2023.
\newblock Gpt-4 technical report.
\newblock \emph{arXiv preprint arXiv:2303.08774}.

\bibitem[{Ouyang et~al.(2022)Ouyang, Wu, Jiang, Almeida, Wainwright, Mishkin, Zhang, Agarwal, Slama, Ray et~al.}]{ouyang2022training}
Long Ouyang, Jeff Wu, Xu~Jiang, Diogo Almeida, Carroll Wainwright, Pamela Mishkin, Chong Zhang, Sandhini Agarwal, Katarina Slama, Alex Ray, et~al. 2022.
\newblock Training language models to follow instructions with human feedback.
\newblock \emph{Advances in Neural Information Processing Systems}, 35:27730--27744.

\bibitem[{Shi et~al.(2024)Shi, Tang, Zhang, Zhang, and Yang}]{shi2024surveyemployinglargelanguage}
Liang Shi, Zhengju Tang, Nan Zhang, Xiaotong Zhang, and Zhi Yang. 2024.
\newblock \href {https://arxiv.org/abs/2407.15186} {A survey on employing large language models for text-to-sql tasks}.
\newblock \emph{Preprint}, arXiv:2407.15186.

\bibitem[{Van Der~Maaten et~al.(2024)}]{maaten2024llama3}
Laurens Van Der~Maaten et~al. 2024.
\newblock The llama 3 herd of models.
\newblock \emph{arXiv preprint arXiv:2407.21783}.

\bibitem[{Wolf et~al.(2019)Wolf, Debut, Sanh, Chaumond, Delangue, Moi, Cistac, Rault, Louf, Funtowicz et~al.}]{wolf2019huggingface}
Thomas Wolf, Lysandre Debut, Victor Sanh, Julien Chaumond, Clement Delangue, Anthony Moi, Pierric Cistac, Tim Rault, R{\'e}mi Louf, Morgan Funtowicz, et~al. 2019.
\newblock Huggingface's transformers: State-of-the-art natural language processing.
\newblock \emph{arXiv preprint arXiv:1910.03771}.

\bibitem[{Xu et~al.(2023)Xu, Sun, Zheng, Geng, Zhao, Feng, Tao, and Jiang}]{xu2023wizardlm}
Can Xu, Qingfeng Sun, Kai Zheng, Xiubo Geng, Pu~Zhao, Jiazhan Feng, Chongyang Tao, and Daxin Jiang. 2023.
\newblock Wizardlm: Empowering large language models to follow complex instructions.
\newblock \emph{arXiv preprint arXiv:2304.12244}.

\bibitem[{Yu et~al.(2018)Yu, Zhang, Yang, Yasunaga, Wang, Li, Ma, Li, Yao, Roman, Zhang, and Radev}]{yu-etal-2018-spider}
Tao Yu, Rui Zhang, Kai Yang, Michihiro Yasunaga, Dongxu Wang, Zifan Li, James Ma, Irene Li, Qingning Yao, Shanelle Roman, Zilin Zhang, and Dragomir Radev. 2018.
\newblock \href {https://doi.org/10.18653/v1/D18-1425} {{S}pider: A large-scale human-labeled dataset for complex and cross-domain semantic parsing and text-to-{SQL} task}.
\newblock In \emph{Proceedings of the 2018 Conference on Empirical Methods in Natural Language Processing}, pages 3911--3921, Brussels, Belgium. Association for Computational Linguistics.

\bibitem[{Zhong et~al.(2024)Zhong, Zhong, Sun, Jin, Qin, and Zhang}]{zhong2024synthet2cgeneratingsyntheticdata}
Ziije Zhong, Linqing Zhong, Zhaoze Sun, Qingyun Jin, Zengchang Qin, and Xiaofan Zhang. 2024.
\newblock \href {https://arxiv.org/abs/2406.10710} {Synthet2c: Generating synthetic data for fine-tuning large language models on the text2cypher task}.
\newblock \emph{Preprint}, arXiv:2406.10710.

\end{thebibliography}

\appendix

\section{Appendix-1}
\section{Schema Generation Process}
\label{apdx_schema_generation}
We start with a seed list of 10 domains (e-commerce, IT Management, finance etc) as well as the domains in the Neo4J example databases on their website \cite{neo4j}. Then we prompt a Mixtral-8\*22B model with higher temperature (0.8) to generate more such domains. Pooled together this yeilds 693 schemas which are split into Train and Test as shown in Table-\ref{tab:dataset_table}.

\subsection{Nodes and Relationships}
\label{apdx_nodes_and_relationships}
We start of by contructing a skeleton schema which includes the nodes and relationships that are \textit{plausible} in the given domain. We elicit responses by conditioning on varying number of nodes and relationships, as well as various query taxonomies to cover a wide range of complexity in the graph as shown in Figure-\ref{fig:skeleton_schema} 

\subsection{Final Schema}
\label{apdx_final_schema}
Once we obtain the nodes and relationships sets, we come up with the full schema along with datatypes, properties and directed edges as shown in Figure-\ref{fig:complete_schema}. We elicit the model to reason through matching the nodes with the generated relationships and obtain a final schema. We manually vet 25\% of the schemas to ensure diversity, coherence and real world usefulness.

\section{Question Generation}
\label{apdx_question_generation}
For every schema, 20 elicit questions at a time from Mixtral-8*22B by sequentially conditioning it on a randomly selected 7 query types. This ensured a diverse question set covering all domains and query types. We pass these questions through an simple LLM validation to ensure they are not too vague, for.e.g \textit{``How many employees report to 'John Doe'?} rather than \textit{How many employees report to a specific manager?} 

\section{Synthetic Ground Truth Generation}
\label{apdx_synthetic_generation}
For each question, we generate a dummy ground truth, which is of the expected structure, data-type and is plausibly true for that question. The prompt for the same is given in Figure-\ref{fig:synthetic_ground_truth} For e.g.

\textbf{Question:} ``What is the total sales in USD for Apples in the California market and who made the most sales?''

\textbf{Dummy answer:} \{``total\_sales\_usd'': 10000, ``employee'': ``John Doe''\}

\section{Database Infilling}
\label{apdx_database_infill}
To fill the database with in such a way that the dummy answer is the right answer for the question, we come up with both positive (relevant to the question, and dummy answer) and negative data points (irrelevant to the question). The prompt is given in Figure-\ref{fig:code_plan_generation}
 and Figure-\ref{fig:python_code_generation}. A full example is given as well.

\section{Cypher Generation}
\label{apdx_cypher_gen}
We do this in four detailed steps so as to give the model ample reasoning and planning tokens. These include
\begin{itemize}
    \item Analysing the user's question - Figure-\ref{fig:cypher_gen_step1}
    \item Identifying the pertinent nodes, relationships, and properties for the question. Figure-\ref{fig:cypher_gen_step2}
    \item Recalling the best practices and coding guidelines for Cypher, including performance concerns. \ref{fig:cypher_gen_step3}
    \item Generating the final Cypher query. \ref{fig:cypher_gen_step4}
\end{itemize}

\begin{figure*}[h!]
  \centering
  \begin{tcolorbox}[colback=white!95!gray, colframe=black, width=\textwidth, arc=2mm, auto outer arc, boxrule=0.5mm, title=Skeleton Schema Generation]
  \lstset{style=wrappedverbatim}
  \begin{lstlisting}
You are an expert in Neo4j databases. You are given a Neo4J database name. Your job is to come up with a possible list of nodes and relationships in the database. The nodes and relationships should be in such a way that they could exist in a real-world scenario based on the database name provided.

Database Name: {database_name}

INSTRUCTIONS:
You need to design {num_nodes} nodes and {num_relationships} relationships that could be present in the database.

Relationships should be in the format of "RELATIONSHIP_NAME", i.e. all uppercase with spaces replaced by underscores.

** The same relationship can be SHARED by different kinds of nodes. So you should design these relationships such that they can connect various pairs of nodes. **

The nodes and releationships should be in such a way that we can ask the following kinds of queries on them:

{taxonomies}

You MUST explain how the queries of the above taxonomies can be used in the context of the nodes and relationships you have provided. 

Return your response as JSON with the following format:
{{
"nodes": {node_examples},
"relationships": {relationship_examples}
}}

Output your result as: 

Explanation: <your explanation here>

Json response: <your json response here>
  \end{lstlisting}
  \end{tcolorbox}
  \caption{Skeleton schema generation step using Mixtral-8*22B}
  \label{fig:skeleton_schema}
\end{figure*}

% <BEGIN>

\vspace{-25mm}
\begin{figure*}[h!]
  \centering
  \begin{tcolorbox}[colback=white!95!gray, colframe=black, width=\textwidth, arc=2mm, auto outer arc, boxrule=0.5mm, title=Complete Schema Generation]
  \lstset{style=wrappedverbatim}
  \begin{lstlisting}
You are an expert in Neo4j databases. You are given a Neo4J database name.
Your job is to come up with a possible schema for nodes and relationships in 
the database.

# Instructions:
- Note that the node and relationship properties can have any of the 
following types:
BOOLEAN, DATE, DURATION, FLOAT, INTEGER, LIST, LOCAL DATETIME, LOCAL TIME,
POINT, STRING,
ZONED DATETIME, and ZONED TIME.

- It is important that the generated schema can be used to create queries 
such as the following:
Taxonomies:
{taxonomies}
The nodes should be formatted as given in the example below.
Example: If the node is 'Person', you should write it as:
Person {{name: STRING, age: INTEGER, date_of_birth: DATETIME}}
The relationship properties should be formatted as given in the example 
below.
Example: If the relationship is 'WORKS_AT', you should write it as:
WORKS_AT {{ employee_id: STRING, since: DATETIME, salary: INTEGER}}
The relationships should be formatted as given in the example below.
Example: If the relationship is 'WORKS_AT', you should write it as:
(:Person)-[:WORKS_AT]->(:Employer)

# Example:
Database Name: movies
NODES: [Movie, Person]
RELATIONSHIPS: [ACTED_IN, REVIEWED, DIRECTED, PRODUCED, WROTE, FOLLOWS]
Schema:
```
Node properties:
Movie {{title: STRING, votes: INTEGER, tagline: STRING, released: INTEGER}}
Person {{born: INTEGER, name: STRING}}

Relationship properties:
ACTED_IN {{roles: LIST}}
REVIEWED {{summary: STRING, rating: INTEGER}}

The relationships:
(:Person)-[:ACTED_IN]->(:Movie)
(:Person)-[:DIRECTED]->(:Movie)
(:Person)-[:PRODUCED]->(:Movie)
(:Person)-[:WROTE]->(:Movie)
(:Person)-[:FOLLOWS]->(:Person)
(:Person)-[:REVIEWED]->(:Movie)
```
# Task:
Database Name: {database_name}
NODES: [{nodes_list}]
RELATIONSHIPS: [{relationships_list}]

Explanation: <Explain how the nodes, releationships, and properties can be used to frame queries 
as per the taxonomies provided>.
Schema:
```
Node properties:
<your node properties here>
Relationship properties:
<your relationship properties here>
The relationships:
<your relationships here>
```
MAKE ABSOLUTELY SURE THAT SCHEMA IS IN THE ABOVE FORMAT WITH ```<schema>``` tags.
  \end{lstlisting}
  \end{tcolorbox}
  \caption{Complete schema generation step using Mixtral-8*22B}
  \label{fig:complete_schema}
\end{figure*}

\begin{figure*}[h!]
  \centering
  \begin{tcolorbox}[colback=white!95!gray, colframe=black, width=\textwidth, arc=2mm, auto outer arc, boxrule=0.5mm, title=Question Generation]
  \lstset{style=wrappedverbatim}
  \begin{lstlisting}
You are an expert in Neo4j databases. I will provide you with a schema, and 
your task is to generate 20 unique questions directly related to that 
specific graph schema.

## Task:
Generate 20 questions that focus on the schema's nodes and relationships.

## Steps for Question Generation:
1. Analyze the Schema: Examine the provided schema and identify relevant 
nodes and relationships.
Select Nodes and Relationships: Based on the query type, choose nodes and 
relationships to form
the questions.
2. Generate Diverse Questions: Create 20 questions, each addressing 
different  aspects of the schema. Ensure no two questions are similar.
3. Cover Key Aspects: Each question should focus on distinct parts 
of the schema, such as relationships between nodes, node properties, or 
node types.
4. Vary Complexity: Ensure the questions range from basic to advanced, 
covering various levels of query complexity.
Random Selection: Randomly select nodes or relationships when forming 
each question, ensuring diversity in the coverage.
5. Specific Values: When generating questions involving values like date, 
time, money, name, or location, use appropriate placeholder values 
(e.g., "2024-01-01" for a date. "John Smith" for name etc). Be creative!
6. Clarity and Relevance: All questions should be clear, unambiguous, and 
reflective of what a human would ask.

Important:
* Ensure each question includes all the information necessary for a 
meaningful answer.
* Generate exactly 20 questions, ensuring they cover different aspects 
of the schema and that
none are repetitive.

Type of query for which questions need to be generated are:
{Query Type}

Schema:
{Schema}

  \end{lstlisting}
  \end{tcolorbox}
  \caption{Question generation step using Mixtral-8*22B}
  \label{fig:question_generation}
\end{figure*}

\begin{figure*}[h!]
  \centering
  \begin{tcolorbox}[colback=white!95!gray, colframe=black, width=\textwidth, arc=2mm, auto outer arc, boxrule=0.5mm, title=Synthetic ground truth generation]
  \lstset{style=wrappedverbatim}
  \begin{lstlisting}
You are an expert in Neo4j databases and creating test data. I have a Neo4j schema and a user query. I am creating a test dataset to validate my Neo4j Cypher queries. Your task is to analyze both the schema and the user question to determine which nodes, fields, and relationships are involved.

Based on your analysis, generate a dummy answer that closely mirrors what would be returned from a Neo4j query, without any post-processing. The fields in the dummy data should directly reflect the schema and be relevant to the user query. 
**
Do not include fields unrelated to the question or absent from the schema.
**
The generated dummy data must:
- Be complete, concise, and accurate.
- Match the format returned by a Neo4j database.
- Use appropriate fields from the schema, without any unnecessary data.
- Reflect counts (votes, followers, etc.) as realistic and generally below 
50, unless otherwise specified.
- Use correct ranges and units for numerical values 
(e.g., convert "1 million" to "1000000").
- Ensure unique values for fields like IDs, timestamps, 
or other attributes that require uniqueness in the database.

The output must be in valid, properly formatted JSON:
```json
{"Answer": <Dummy ANSWER>}
```
Before generating the answer, clearly define the nodes and relationships essential for covering the user question. If there are multiple records in the dummy data, ensure unique values for attributes such as IDs, timestamps, and steps.

Example user question:  
Which Disney character laughed how many times, and what is their favorite color?

```json
{
  "Answer": [
    {
      "characterid": "b92",
      "characterName": "Mini",
      "laughed": 100,
      "favorite_color": "Red"
   }, 
   {
      "characterid": "d989", 
      "characterName": "Jimmi", 
      "laughed": 10, 
      "favorite_color": "Blue"}]}
```
Schema:
{SCHEMA}

User Question:
{USER_QUESTION}  
\end{lstlisting}
  \end{tcolorbox}
  \caption{Synthetic ground truth generation step using Mixtral-8*22B}
  \label{fig:synthetic_ground_truth}
\end{figure*}

\begin{figure*}[h!]
  \centering
  \begin{tcolorbox}[colback=white!95!gray, colframe=black, width=\textwidth, arc=2mm, auto outer arc, boxrule=0.5mm, title=Code plan generation for database infilling]
  \lstset{style=wrappedverbatim}
  \begin{lstlisting}
You are an expert in writing Python code. I am working on creating test data to validate Neo4j Cypher queries. 
You will be provided with the Neo4j schema, a user question, and a ground truth answer. 
Your task is to generate test case data that will populate an empty Neo4j database. 

This will allow me to check if the Cypher query, based on the user's question, retrieves the correct result as per the ground truth answer. To ensure robust validation, the data you create must return the exact ground truth answer when queried, but the database should also include additional "negative" data points. These negative points must not interfere with the correct answer and will test the accuracy of the query. 

Follow these steps:
Steps:
1. Analyze the User Question and Schema: Identify relevant nodes, relationships, and fields based on the schema and user question. Understand which entities are crucial to construct the ground truth answer.
2. Plan Data Population: Develop a structured plan that describes how the data will be populated. Include both the ground truth data and additional negative data points.
3. Write Cypher Queries:  
   Provide the exact Cypher queries for:
   - Creating nodes and relationships for the ground truth answer.
   - Creating negative data points that do not match the answer but help ensure the test is comprehensive.
4. Comprehensive Negative Data: For the negative data points, ensure the information is random, and distinctly different from the ground truth. Include details like names, summaries, and other fields, making sure the negative data does not overlap with the ground truth.
5. Limit Negative Data Points: Do not create more than 5 negative data points. This ensures that the negative data is limited and doesn't overwhelm the test case.
6. Unique Fields for Negative Data: Fields like IDs, names, locations, or titles should be unique, especially in negative data points. Specify which fields require unique values, using UUIDs or similar approaches. Ensure this applies only to negative data; the ground truth must not use UUIDs.
7. UUID Usage in Negative Data: Assign UUIDs to variables before using them in the queries for negative data. 
8. Relationship Creation: Create relationships between nodes using their IDs. Use the `MATCH` statement before creating relationships to ensure that the nodes exist and the correct connections are established.
9. Correct Range of Values: When populating fields like money, votes, or similar data, ensure they align with the question. For example, if the question mentions 1 million, use 1000000; for 1.2 million, use 1200000.

Key Points to Remember:
- Order of Execution: Ensure that nodes are created before relationships. 
- Use `MATCH` to verify node existence before establishing relationships.
- Correctness of Identifiers: Double-check that identifiers (like IDs) match between node creation and relationship creation queries. For instance, if a fruit node is created with `id = fruit1`; the same ID should be used in relationships.

For example:
```cypher
// Example for creating nodes and relationships
CREATE (fruit:Fruit {id: 'fruit1', name: 'apple'});
CREATE (juice:Juice {id: 'juice1', name: 'apple juice'});
MATCH (fruit:Fruit {id: 'fruit1'})
MATCH (juice:Juice {id: 'juice1'})
CREATE (fruit)-[:JUICED]->(juice);
```
- Ground Truth Accuracy: The ground truth answer must be present in the data. 
This ensures the test works as expected, and only the ground truth will produce a valid answer.
- Proper Relationship Creation: Ensure relationships are established correctly by matching node IDs before creating the relationship.

Schema:
{SCHEMA}

User Question:
{USER_QUESTION}

Ground Truth Answer:
{SYNTHETIC_ANSWER_RESPONSE}
\end{lstlisting}
  \end{tcolorbox}

  \caption{Code plan generation step using Gpt-4}
  \label{fig:code_plan_generation}
\end{figure*}

\begin{figure*}[h!]
  \centering
  \begin{tcolorbox}[colback=white!95!gray, colframe=black, width=\textwidth, arc=2mm, auto outer arc, boxrule=0.5mm, title=Python code generation for database infilling]
  \lstset{style=wrappedverbatim}
  \begin{lstlisting}
You are an expert in writing Python code. I am developing a test set of data to verify Neo4j Cypher queries. I will provide the Neo4j schema, user question, ground truth answer, and a code plan. Your task is to create test case data that populates an empty Neo4j database so that when 
Cypher is executed based on the user's question, it returns the correct answer from the database.

The data must ensure that querying the DB returns only the ground truth answer for the given question. Additionally, the database should contain negative data points that do not satisfy the query, ensuring the robustness of the test case. Follow these steps carefully:

Steps:
1. Analyze the schema and user question: Identify relevant nodes, fields, and relationships needed to answer the question.
2. Refer to the code plan: Follow the provided plan for structuring the data generation code.
3. Create relationships and nodes: Ensure all required relationships and nodes are generated in the database.
4. Write the Python code in a function `create_data()`: Return a list of Cypher queries that populate the DB to support the query validation.
5. No execution logic required: The function should return only the list of queries, not execute them.
6. Use real timestamps: Any fields like timestamps must reflect actual values.
7. Ground truth must satisfy the query: Ensure that only the ground truth data satisfies all conditions, and negative data does not.
8. Generate up to 5 negative data points: Each negative example should differ entirely from the ground truth (e.g., UUIDs, random names, summaries). Ensure negative data points are not more than five.
9. Use `MATCH` to ensure relationship correctness: Ensure relationships are created by matching node IDs before defining relationships.

Code Writing Suggestions:
- Avoid errors with f-strings by using string concatenation or `.format()` when needed.
- Assign UUIDs to variables before using them in queries to prevent errors.
- When creating relationships, first use `MATCH` to ensure nodes exist, then define the 
relationships by their node IDs.

Key Points:
- Order of execution: Ensure nodes exist before creating relationships.
- Correctness of identifiers: Verify that `MATCH` statements correctly reference nodes created earlier.
- Consistency: Ensure the actual answer data perfectly satisfies the question, and negative examples do not match the query.

Example:
```
// Create fruit and juice nodes
CREATE (fruit:Fruit {id: 'fruit1', name: 'apple'});
CREATE (juice:Juice {id: 'juice1', name: 'apple juice'});

// Create the "Juiced" relationship
MATCH (fruit:Fruit {id: 'fruit1'})
MATCH (juice:Juice {id: 'juice1'})
CREATE (fruit)-[:JUICED]->(juice);
```
Important:
- Relationships: Ensure relationships are properly created by matching node IDs first.
- Correctness: Only the ground truth data should match the query conditions. Negative data should 
never fulfill the query.
- Return format: Return the Python code wrapped in ```python  ``` tags.

Schema:  
{SCHEMA}
User Question:  
{USER_QUESTION}
Ground Truth Answer:  
{SYNTHETIC_ANSWER_RESPONSE}
Code Plan:  
{CODE_PLAN}
  \end{lstlisting}
  \end{tcolorbox}
  \caption{Python code generation step using Gpt-4}
  \label{fig:python_code_generation}
\end{figure*}

\begin{figure*}[h!]
  \centering
  \begin{tcolorbox}[colback=white!95!gray, colframe=black, width=\textwidth, arc=2mm, auto outer arc, boxrule=0.5mm, title=Cypher generation - Analyse user request]
  \lstset{style=wrappedverbatim}
  \begin{lstlisting}
You are helpful and expert Neo4j and generating Cypher queries assistant.
You will be given
- Neo4j schema
- User question related to the given schema
    
<neo4jschema>
    {SCHEMA}
</neo4jschema>

<question>
    {USER QUESTION}
</question>

YOUR INSTRUCTIONS:-
You are a Neo4j expert. Follow these STEP BY STEP:

1. **Identify Nodes and Relationships**:

  - Examine the schema to identify the different types of nodes (entities) and relationships 
  (edges) between them.

2. **Node Properties**:
  - Note the properties (attributes) of each node type.

3. **Relationship Properties**:
  - Note the properties of each relationship type.

4. **Indexes and Constraints**:
  - Check for any indexes or constraints that might be relevant for query optimization.

5. **Break Down User Question**:
  - Analyze the user's question step by step, using the provided schema as grounding.
  - Understand what the user needs, keeping in mind the eventual answer.
  - For units like 1 million or 1 dozen, convert them to their base forms (e.g., 1 million to 
  1000000, 1 dozen to 12) when generating Cypher queries.

6. **Formulate the Response**:
  - Use the identified nodes, relationships, and their properties to inform your understanding of 
  the user's question.
  - Ensure that any indexes and constraints are considered when formulating your response.
  - Formulate a clear breakdown of the user question and the analysis of the schema.
DO NOT GENERATE THE CYPHER QUERY, JUST FOLLOW THE GIVEN INSTRUCTIONS!
\end{lstlisting}
  \end{tcolorbox}
  \caption{Cypher generation: Analyse question step using Gpt-4}
  \label{fig:cypher_gen_step1}
\end{figure*}

\begin{figure*}[h!]
  \centering
  \begin{tcolorbox}[colback=white!95!gray, colframe=black, width=\textwidth, arc=2mm, auto outer arc, boxrule=0.5mm, title=Cypher generation - Relate user request to schema]
  \lstset{style=wrappedverbatim}
  \begin{lstlisting}
You are helpful and expert Neo4j and generating Cypher queries assistant.
You will be given
- Neo4j schema
- User question related to the given schema
- Analysis of the Neo4j schema, the nodes and the relationships, entities between them, and the 
user question.

<neo4jschema>
{SCHEMA}
</neo4jschema>

<question>
{USER QUESTION}
</question>

<schema_and_question_analysis>
{STEP 0 RESPONSE}
</schema_and_question_analysis>

YOUR INSTRUCTIONS:-
Follow these step by step: 
1. Identify which nodes (entities) from the given schema are important in answering the user 
question and forming the correct Cypher query. Keep track of these nodes. Whenever any kind of ID 
is present in a node, make sure to add it so the final answer includes it along with other 
important properties needed to answer the question. Do not make up any properties that are not 
present in the schema.
2. For all identified important nodes, list all relationships related to those nodes and entities 
individually. Do not create imaginary relationships; only consider the relationships that are 
present in the schema.
3. For all identified important nodes and relationships, list and filter all properties related 
to those nodes and entities individually. Do not create imaginary properties; only consider the 
properties that are present in the schema. Whenever any kind of ID is present in a node, make 
sure to add it as a property so the final answer includes it along with other important 
properties needed to answer the question. Do not make up any properties that are not present in 
the schema.
4. Identify and filter out only the nodes, relationships, and properties which are important and 
relevant to answering the user's question and creating the correct Cypher query, given the 
schema. List out all the important nodes, relationships, and properties that are required to 
answer the user's question in the end. Whenever any kind of ID is present in a node, make sure to 
add it as a property so the final answer includes it along with other important properties needed 
to answer the question. Do not make up any properties that are not present in the schema.

DO NOT GENERATE THE CYPHER QUERY, JUST FOLLOW THE GIVEN INSTRUCTIONS!
  \end{lstlisting}
  \end{tcolorbox}
  \caption{Cypher generation: Relate user request to the schema step using Gpt-4}
  \label{fig:cypher_gen_step2}
\end{figure*}

\begin{figure*}[h!]
  \centering
  \begin{tcolorbox}[colback=white!95!gray, colframe=black, width=\textwidth, arc=2mm, auto outer arc, boxrule=0.5mm, title=Cypher generation - Incorporate Cypher best practices]
  \lstset{style=wrappedverbatim}
  \begin{lstlisting}
You are helpful and expert Neo4j and generating Cypher queries assistant.
You will be given 1) Neo4j schema 2) User question related to the given schema 3) Analysis of the Neo4j schema, the nodes and the relationships, entities between them, and the user question.
- Filtered list of nodes, relationships, and properties which are important and relevant to 
answering the user's question.
<neo4jschema>
{SCHEMA}
</neo4jschema>
<question>
{USER QUESTION}
</question>
<schema_and_question_analysis>
{STEP 0 RESPONSE}
</schema_and_question_analysis>
<important_nodes_relationships_properties>
{STEP 1 RESPONSE}
</important_nodes_relationships_properties>

YOUR INSTRUCTIONS:-
STRICTLY FOLLOWING THE GIVEN INFORMATION from <filtered_nodes_relationships_properties> and <convoluted_relationships>, think step by step out loud and create a explicit and detailed verbose STEP BY STEP "Cypher generation plan" for how a cypher query can be formulated to achieve what the user wants.
Make sure to explicitly mention nodes, relationships, conditions in your plan.
You MUST NOT WRITE CYPHER STATEMENTS, but instead verbally step by step generate a plan, which will help in forming the correct Cypher query.
During question analysis, for entites with shortforms, for example 1 million, or 1 dozen. Represent them in numbers, for e.g. 1000000 instead of 1 million and 12 instead of dozen.

Additionally, consider all of the following:
-- conditions which are required to filter the identified nodes and relationships (WHERE)
-- aggregation functions (COUNT, SUM, AVG, MIN, MAX, COLLECT, STDDEV, VARIANCE, PERCENTILE_CONT, 
PERCENTILE_DISC, MODE, MEDIAN, ARRAY_AGG)
-- ordering (ORDER BY ASC, ORDER BY DESC)
-- limits (LIMIT, SKIP)
-- return statement, what should be returned. Avoid aggregation with RETURN statements. (RETURN, 
DISTINCT, CASE, apoc.do.when)
-- matching patterns (MATCH, OPTIONAL MATCH)
-- creating nodes and relationships (CREATE, MERGE)
[Truncated]
-- conditional operations (CASE, FOREACH, WITH, apoc.do.when)
-- union operations (UNION, UNION ALL)
-- handling indexes and constraints (CREATE INDEX, CREATE CONSTRAINT, DROP INDEX, 
DROP CONSTRAINT)
-- full-text search (CALL db.index.fulltext.queryNodes, CALL 
db.index.fulltext.queryRelationships)
-- pagination (SKIP, LIMIT)
-- handling transactions (BEGIN, COMMIT, ROLLBACK)

ADDITIONAL CYPHER PRACTICES YOU MUST FOLLOW STRICTLY, so make sure this is followed in you cypher 
generation plan:-
[Truncated.]
  \end{lstlisting}
  \end{tcolorbox}
  \caption{Cypher generation: Incorporate Cypher best practices step using Gpt-4}
  \label{fig:cypher_gen_step3}
\end{figure*}

\begin{figure*}[h!]
  \centering
  \begin{tcolorbox}[colback=white!95!gray, colframe=black, width=\textwidth, arc=2mm, auto outer arc, boxrule=0.5mm, title=Cypher generation: Final cypher generation]
  \lstset{style=wrappedverbatim}
  \begin{lstlisting}
You are helpful and expert Neo4j and generating Cypher queries assistant.
You will be given
- Neo4j schema
- User question related to the given schema
- Analysis of the Neo4j schema, the nodes and the relationships, entities between them, and the user question.
- Filtered list of nodes, relationships, and properties which are important and relevant to answering the user's question.
- A comprehensive Cypher generation plan, which will help you in forming the correct Cypher query.

<neo4jschema>
{SCHEMA}
</neo4jschema>

<question>
{USER QUESTION}
</question>

<schema_and_question_analysis>
{STEP 0 RESPONSE}
</schema_and_question_analysis>

<important_nodes_relationships_properties>
{STEP 1 RESPONSE}
</important_nodes_relationships_properties>

<cypher_generation_plan>
{STEP 2 RESPONSE}
</cypher_generation_plan>

YOUR INSTRUCTIONS:-
STRICTLY FOLLOWING THE GIVEN 'cypher_generation_plan' and other gathered given knowledge about required nodes and relationships, your task is to write me a detailed brief on the plan in way like what is question asking, what are the important details in schema, and other relevant info (Assume I don't have access to the plan so I will be relying on your writeup) and explain how you will generate the cypher then generate the syntactically correct final cypher query, which will give the desired result, in ansering the user's question.

Generated Cypher should be surrounded by ```cypher```. 
For entites with shortforms, for example 1 million, or 1 dozen. Represent them in numbers, for e.g. 1000000 instead of 1 million and 12 instead of dozen.
  \end{lstlisting}
  \end{tcolorbox}
  \caption{Cypher generation: Final cypher generation step using Gpt-4}
  \label{fig:cypher_gen_step4}
\end{figure*}

\begin{figure*}[h!]
  \centering
  \begin{tcolorbox}[colback=white!95!gray, colframe=black, width=\textwidth, arc=2mm, auto outer arc, boxrule=0.5mm, title=Execution Match - LLM-as-Judge]
  \lstset{style=wrappedverbatim}
  \begin{lstlisting}
As an AI model, your task is to evaluate the student's answer based on the given question and the correct answer. The student's answer may not contain all the fields mentioned in the correct answer or vice versa, but it should address the specific elements asked in the question. If the main elements asked in the question are correctly answered, consider it correct. 

For example:
    Example_Question: "Which employees earn more than 40K in salary that live in USA?"
    Example_Correct_Answer: [{{'name': 'John', 'employee_id': 1234, 'salary': 45K, 'country': 'USA'}}, {{'name': 'Adam', 'employee_id': 2763, 'salary': 90K, 'country': 'USA'}}]
    Example_Student_Answer: [{{'emmployee_name': 'Adam'}}, {{'employee_name': 'John'}}]
In this example, student's answer is CORRECT because the question asks for employee and the student gave the employee names (which uniquely determine the employees). And all the values match. So the student's answer is correct.

For example:
    Example_Question: "Which employees earn more than 40K in salary that live in USA?"
    Example_Correct_Answer: [{{'name': 'John', 'employee_id': 1234, 'salary': 45K, 'country': 'USA'}}, {{'name': 'Adam', 'employee_id': 2763, 'salary': 90K, 'country': 'USA'}}]
    Example_Student_Answer: [{{'emmployee_name': 'Adam'}}, {{'employee_name': 'John'}}, {{'employee_name': 'Victor'}}]
In this example, student's answer is INCORRECT because although the student gave the requested items, i.e, employee name, there is an additional value "Victor" which is incorrect. 

Question:
{task}

Correct Answer:
{ground_truth}

Student's Answer:
{predicted}

Think step by step and finally return your final answer as: FINAL_ANSWER: CORRECT/INCORRECT

  \end{lstlisting}
  \end{tcolorbox}
  \caption{Execution Match - LLM-as-Judge}
  \label{fig:execution_match}
\end{figure*}

\end{document}